\documentclass[letterpaper]{article} 
\usepackage{egbib}  
\usepackage{times}  
\usepackage{helvet}  
\usepackage{courier}  
\usepackage[hyphens]{url}  
\usepackage{graphicx} 
\usepackage{natbib}  
\usepackage{caption} 
\usepackage{algorithm}
\usepackage{romanbar}
\usepackage{algorithmic}
\usepackage{times}
\usepackage{subfig}
\usepackage{epsfig}
\usepackage{dsfont}
\usepackage{graphicx}
\usepackage{amsmath}
\usepackage{amssymb}
\usepackage{multirow}
\usepackage{bbding}
\usepackage{pifont}
\usepackage{tablefootnote}
\usepackage{microtype}
\usepackage[table,xcdraw, dvipsnames]{xcolor}
\usepackage{tabularx, booktabs, nicematrix}
\newcolumntype{Y}{>{\centering\arraybackslash}X}

\usepackage{newfloat}
\usepackage{listings}
\DeclareCaptionStyle{ruled}{labelfont=normalfont,labelsep=colon,strut=off} 
\floatstyle{ruled}
\newfloat{listing}{tb}{lst}{}
\floatname{listing}{Listing}
\title{R\textsuperscript{2}Det: Redemption from Range-view Representation for Accurate 3D Object Detection}
\author {
    Yihan Wang\textsuperscript{\rm 1},
    Qiao Yan\textsuperscript{\rm 1},
    Yi Wang\textsuperscript{\rm 2}
}
\affiliations {
    \textsuperscript{\rm 1}Nanyang Technological University\\
    \textsuperscript{\rm 2}The Hong Kong Polytechnic University\\
    WANG1517@e.ntu.edu.sg, QIAO003@e.ntu.edu.sg, yi-eie.wang@polyu.edu.hk
}

\usepackage{bibentry}

\begin{document}

\maketitle

\begin{abstract}
LiDAR-based 3D object detection is of paramount importance for autonomous driving. Recent trends show a remarkable improvement for bird's-eye-view (BEV) based and point-based methods as they demonstrate superior performance compared to range-view counterparts. This paper presents an insight that leverages range-view representation to enhance 3D points for accurate 3D object detection. Specifically, we introduce a Redemption from Range-view Module (R\textsuperscript{2}M), a plug-and-play approach for 3D surface texture enhancement from the 2D range view to the 3D point view. R\textsuperscript{2}M comprises BasicBlock for 2D feature extraction, Hierarchical-dilated (HD) Meta Kernel for expanding the 3D receptive field, and Feature Points Redemption (FPR) for recovering 3D surface texture information. R\textsuperscript{2}M can be seamlessly integrated into state-of-the-art LiDAR-based 3D object detectors as preprocessing and achieve appealing improvement, e.g., $1.39\%$, $1.67\%$, and $1.97\%$ mAP improvement on easy, moderate, and hard difficulty level of KITTI \textit{val} set, respectively. Based on R\textsuperscript{2}M, we further propose R\textsuperscript{2}Detector (R\textsuperscript{2}Det) with the Synchronous-Grid RoI Pooling for accurate box refinement. R\textsuperscript{2}Det outperforms existing range-view-based methods by a significant margin on both the KITTI benchmark and the Waymo Open Dataset. Codes will be made publicly available.
\end{abstract}

\section{Introduction}\label{sec_intro}

\begin{figure}[t]
    \centering
    \includegraphics[width=1\linewidth]{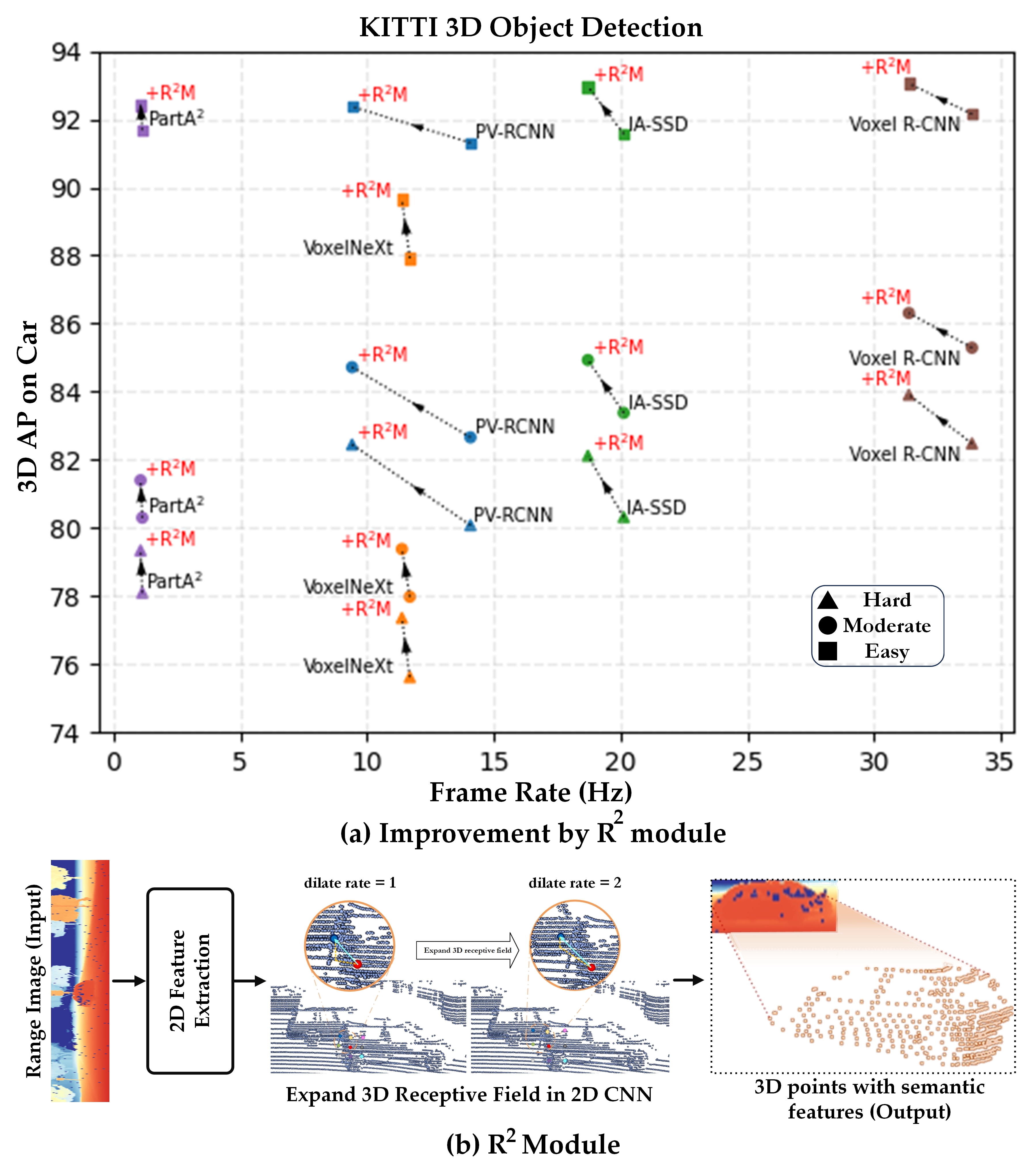}
    \vspace{-5mm}
    \caption{(a) Our proposed plug-and-play R\textsuperscript{2} Module improves the performance of existing 3D object detectors at a small cost of inference speed. (b) The sketch of R\textsuperscript{2} Module. It encodes the input range images with 2D features by expanding the 3D receptive field and outputs 3D points with semantic features.}
    \label{fig:intro}
    \vspace{-4mm}
\end{figure}

The LiDAR sensor plays a fundamental role in enabling the perception system for autonomous driving. However, the current scene understanding methods heavily rely on expensive 3D point cloud data generated by LiDAR sensors. To address this cost challenge, emerging LiDAR sensors that output range images have rapidly gained notice in recent years \cite{rsn}. These sensors offer a significant cost advantage, nearly four times cheaper than LiDAR systems generating point clouds. Additionally, range image-based LiDAR sensors establish a perfect 1-to-1 relationship between a pixel and a 3D point without discrete or resampling steps involved \cite{ouster}. Due to this clear relationship between pixels and 3D points, those LiDAR sensors hold great potential for providing precise 3D semantic information. Motivated by this potential, several related works on semantic scene segmentation have been developed, such as \cite{rangenet++,squeezeseg,rangeseg,randla,transrvnet}.

However, the potential of pixel-wise semantics has yet to be fully harnessed in the task of 3D object detection. Despite the attempts by existing 3D point-wised methods \cite{pointnet,pointnet++, ia-ssd,dtssd, Chen2022SASASS} to alleviate computational burdens through set abstraction or downsampling modules, these methods still face challenges in terms of efficiency for practical implementation. On the other hand, there is another kind of methods that rely on BEV (or voxels) \cite{votr,octr,voxset,voxenext}, offering faster inference times, but they still struggle to achieve satisfactory levels of accuracy. Point-view and BEV methods have received thorough investigation, but range-view methods remain relatively underexplored. The information provided by range-view representation has not been fully leveraged in existing approaches.

As previously highlighted, range images establish a direct correspondence between pixels and their corresponding 3D points. This implies a more effective acquisition of semantic information for each 3D point from range images, i.e., extracting features within the 2D plane and subsequently translating them back to 3D points. This is an advantage over conventional 3D point-wise feature extraction techniques in terms of computational complexity. However, existing range-view-based 3D detectors \cite{rangeioudet,rangedet,lasernet,RCD} process range images directly based on 2D CNN backbones. While efficient, they overlook a crucial issue, the loss of 3D surface texture information prevalent in 2D range images. This particular information encapsulates the interrelations among points within the point clouds. For instance, as depicted in Fig.~\ref{fig:intro}(b), while range images uphold the relative spatial distances between points, they unavoidably omit \emph{all} the 3D surface information, which fundamentally contributes to the intricate relationships among points in 3D coordinates.

In this paper, we address the abovementioned challenges through a meticulously designed Redemption from Range-view Module (R\textsuperscript{2}M), a plug-and-play manner to extract 3D surface textures from 2D range images, as shown in Fig.~\ref{fig:intro}(b). First, R\textsuperscript{2}M comprises a BasicBlock and a novel Hierarchical-dilated (HD) Meta Kernel. This combination effectively enhances 2D range-view point-wise semantic features by addressing variations in object sizes and significantly expanding the 3D receptive field. Second, to counter the loss of 3D surface texture information inherent in range images, we introduce the Feature Points Redemption (FPR). The FPR retrieves the lost information and transforms semantic feature points into 3D space, thereby reinstating object surface textures and minimizing the impact of the 3D information loss inherent in 2D range-view representation. Importantly, this feature extraction operation is considerably more efficient than existing 3D point-wise extraction methods. The 3D points generated by the R\textsuperscript{2}M inherently contain semantic information. As a result, these points can be seamlessly integrated into the input stage of existing single-modality-based 3D object detectors, enhancing their input data and subsequently improving performance on all three difficulty levels of KITTI \textit{val} set, as shown in Fig. \ref{fig:intro}(a). Moreover, we build the R\textsuperscript{2}Detector (R\textsuperscript{2}Det) with the proposed Synchronous-Grid (S-Grid) RoI Pooling strategy for box refinement, which uses two different grid sizes in parallel to aggregate region-of-interest (RoI) features from each proposal. This multi-sampling approach facilitates the abstraction of more contextual features, leading to more accurate box prediction. 

Our contributions can be summarized as follows. (1) We propose an efficient plug-and-play R\textsuperscript{2}M, which enhances the 2D feature extraction by expanding the 3D receptive field and provides an elegant strategy for resolving the 3D surface texture loss and exploring the point-wise semantics of range images. (2) We introduce R\textsuperscript{2}Det, a novel approach that addresses the limitations of range-view representation and enables more accurate 3D object detection. (3) We perform exhaustive trials on both the KITTI dataset \cite{kitti} and the Waymo Open Dataset \cite{Waymo}. As a result, our proposed R\textsuperscript{2} module can boost 3D object detectors consistently with small frame rate sacrifice (see Table \ref{tab:effect_of_R2}). Our proposed R\textsuperscript{2}Det outperforms existing range-view-based methods significantly and achieves state-of-the-art performance on both datasets. 

\begin{table}[h]
\centering
    \setlength{\tabcolsep}{6pt} 
    \renewcommand{\arraystretch}{0.8} 
\begin{tabular}{cccc}
\toprule
 & \multicolumn{3}{c}{AP$_{\textbf{3D}}$(\%)} \\
\multirow{-2}{*}{Method} & Easy & Mod. & Hard \\ \hline
PV-RCNN \cite{pvrcnn}& 91.31 & 82.66 & 80.08 \\
R\textsuperscript{2}M+PV-RCNN & 92.37 & 84.72 & 82.45 \\ 
\rowcolor[HTML]{96FFFB} 
\textit{Improvement} & \textit{+1.06} & \textit{+2.06} &\textit{+2.37} \\ \hline
VoxelNeXt\cite{voxenext} & 87.92 & 77.98 & 75.62 \\
R\textsuperscript{2}M+VoxelNeXt& 89.65 & 79.38 & 77.36 \\ 
\rowcolor[HTML]{96FFFB} 
\textit{Improvement} & \textit{+1.73} & \textit{+1.40} &\textit{+1.73} \\ \hline 
IA-SSD\cite{ia-ssd} & 91.57 & 83.38 & 80.32 \\
R\textsuperscript{2}M+IA-SSD& 92.96 & 84.93 & 82.12 \\ 
\rowcolor[HTML]{96FFFB} 
\textit{Improvement} & \textit{+1.39} & \textit{+1.55} &\textit{+1.80} \\ \bottomrule
\end{tabular}
\vspace{-3mm}
\caption{Effect of the R\textsuperscript{2} Module on the KITTI \textit{val} split with AP calculated by 40 recall positions.}
\label{tab:effect_of_R2}
\end{table}

\begin{figure*}[!ht]
    \centering
    \includegraphics[width=0.99\linewidth]{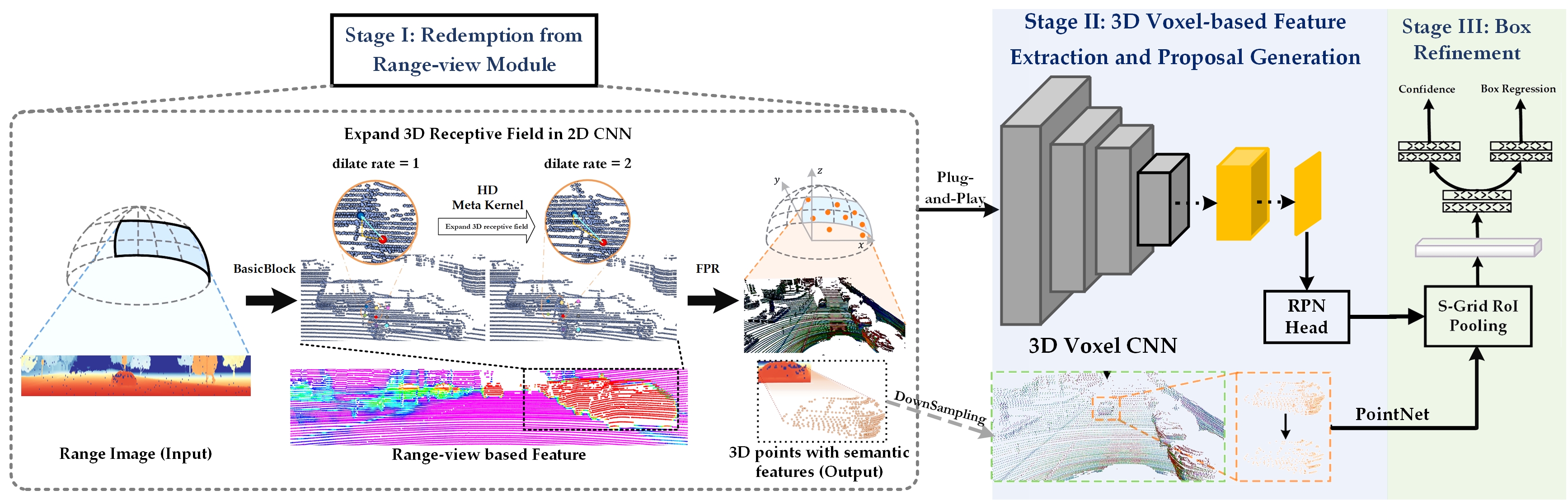}
    \vspace{-3mm}
    \caption{Architecture of our proposed R\textsuperscript{2}Det. The range images are initially processed through a BasicBlock and the HD Meta Kernel. The FPR module transforms these images from 2D semantic pixels to 3D points with surface texture features. Then 3D feature points are voxelized and fed into a 3D sparse convolution-based backbone. This backbone encoders multi-scale semantic features and generates 3D proposals. Meanwhile, the redeemed feature points from range images are downsampled. Finally, these downsampled features and candidate boxes are passed through the S-Grid RoI Pooling to refine the candidate boxes.}
    \label{fig:Architecture}
    \vspace{-2mm}
\end{figure*}

\section{Related Work}\label{sec_related_work}
\textbf{3D Object Detection on Point Clouds.} Point clouds remain a favored input representation due to their accurate depth information and resilience to various weather conditions. However, the substantial volume of unstructured data poses an initial challenge in determining the appropriate input data representation. Some efforts have been made to project point clouds into multiple views \cite{AVOD,MV3D,MVF,DeepContinuousFusion,hdnet}, relying on manual feature extraction. Continuous works \cite{voxelnet, pointpillars,voxelrcnn, voxset,votr,octr,gd-mae} have replaced the manual step with a learnable end-to-end voxel feature abstraction module. These voxelization-based methods offer higher efficiency but often suffer from unsatisfactory accuracies.

Contrasting with approaches relying on the hybrid view and voxels, PV-based methods establish object identification relative to individual points. Initial works like \cite{F-PointNet,F-Convnet,IPOD} integrate \cite{pointnet, pointnet++} into their frameworks. PointRCNN \cite{pointrcnn} introduces a revolutionary box refinement module. Despite considerable accuracy improvements, these methods still grapple with latency concerns. 3DSSD \cite{3dssd} employs a set abstraction layer to downsample point clouds and mitigate latency. Similarly, SASA \cite{Chen2022SASASS} advances latency reduction by introducing an innovative set abstraction module for foreground point segmentation. DTSSD \cite{dtssd} proposes a central density-aware enhancement module to address latency. Further latency reduction remains an ongoing challenge.

\textbf{3D Object Detection on Range Images.} The solution for transplanting the fully convolutional network onto the RV of point clouds is investigated by \cite{VehicleDF}. Based on this approach, LaserNet \cite{lasernet} predicts the probability of bounding boxes for representing the uncertainty of detection using 2D CNNs. \cite{RCD} similarly uses 2D CNNs while it novelly presents a dynamic adjustable dilate rate mechanism for adapting the scale variation caused by range images. In comparison to \cite{RCD}, \cite{rangercnn} and its continuous work \cite{rangeioudet} extract RV feature maps with multiple dilate rates. \cite{rsn} proposes to regress 3D boxes by transferring foreground points from the RV to the PV. Innovatively, RangeDet \cite{rangedet} exploits point cloud geometric information from RV for feature extraction.

\section{Methodology}\label{sec:RVFE}

\subsection{Overall Architecture}
Figure~\ref{fig:Architecture} presents the schematic for the entire structure of R\textsuperscript{2}Det. The input to the system is a five-channel range view of a point cloud (also known as a range image), which includes x, y, z, intensity, and range data. The range image is firstly encoded by a BasicBlock \cite{basicblock}, then the output feature maps are stacked into Hierarchical-dilated (HD) Meta Kernel. The output 2D feature map is transferred from the range view to the point view in the Feature Points Redemption (FPR) module. These steps form Stage I: R\textsuperscript{2}M. Afterward, these 3D feature points are split into two streams. One is partitioned into voxels and fed into 3D voxel CNNs for proposal generation, and the other is downsampled into a set of keypoints. Finally, the generated candidate boxes and the downsampled keypoints are fed into the Synchronous-Grid (S-Grid) RoI pooling module  for box refinement.

\subsection{Stage I: Redemption from Range-view Module} \label{sec:retrieve}
\textbf{Range-view Feature Extraction.} Benefiting from incorporating 3D geometric information into 2D CNN, \textit{Meta-kernel} convolutional layer proposed in \cite{rangedet} outperformed conventional 2D CNNs. Inspired by this, we introduce a straightforward yet impactful module, Hierarchical Dilated (HD) Meta-kernel. This module enhances the utilization of 3D geometric information within 2D CNNs, leading to improved performance. Additionally, our approach innovatively extends the 3D receptive field in 2D CNNs. This extension allows for the dynamic generation of adapted weights from related 3D coordinates, addressing scale variation challenges in range images. 

Similar to \textit{Meta-kernel} \cite{rangedet}, the HD Meta Kernel can also be decomposed into the feature sampler, the element-wise processor, and the feature accumulator. Different from \textit{Meta-kernel}, we incorporate another dilate rate branch in the feature sampler, and the feature accumulator also differs. As a result, the feature sampler can be formularized as two kernels $\mathcal{K}_1$ and $\mathcal{K}_2$: 
\begin{equation}
    \mathcal{K}_1 = \{(d^1_h, d^1_w)| \, d^1_h, d^1_w \in [-1, 1], d^1_h, d^1_w \in \mathbb{Z} \}
    \label{eqn:standconvkernel}
\end{equation}
\begin{equation}
    \mathcal{K}_2 = \{(d^2_h, d^2_w)| \, d^2_h = 2d^1_h, d^2_w= 2d^1_w\}
    \label{eqn:dilateconvkernel}
\end{equation}
And the feature accumulator can be described as follows:
\begin{align}
    O_{l_o} = Concat(\mathbf{FC}(Concat&(I_{l_o \Leftrightarrow l_n})), \notag \\
    &\mathbf{FC}(Concat(I_{l_o' \Leftrightarrow l_n'})))
    \label{eqn:finalconcat}
\end{align}
where $o, o'$ define the sampling site in each stream and $n, n' = \{0,1,...,7, 8\}$ are the corresponding nearest eight neighbors,  $I_{l_o \Leftrightarrow l_n}$ denotes the output features from the element-wise processor, $\mathbf{FC}$ and $Concat$ represent the fully connected layer and the concatenating process respectively. 

\textbf{Feature Points Redemption.} In the FPR step, range-view feature points are transformed into point-view representation and subsequently processed by 3D CNNs. The one-to-one mapping $\Pi: \mathbb{R}^2 \mapsto \mathbb{R}^3$ between each pixel $(u,v)$ at the range image and each point $p=(x,y,z)$ is established as Eq.\eqref{eq_fpr}: 
\begin{equation}
\label{eq_fpr}
\begin{gathered}
\begin{aligned}
x &= \sqrt{\frac{r^2 \{ 1 - \sin^2{ ((1-v/h)\mathbf{F}-\mathbf{F}_{up})  } \}}{\sec^2{\{(1-2u/w)/\pi\}}}} \\
y &= x \cdot \tan{\{(1-2u/w)/\pi\}}\\
z &= \sin{\{ (1-v/h)\mathbf{F} - \mathbf{F}_{up}\}} r
\end{aligned}
\end{gathered}
\end{equation}
where $\mathbf{F}=\mathbf{F}_{up}+\mathbf{F}_{down}$ is the vertical field-of-view, $(h,w)$ are the height and width of the projected range image, $r$ is the range value of each pixel. Thus, each pixel $I_{(u,v)}=f^{(64)}$ carrying with a feature embedding of surface texture information are projected back into the corresponding point $\mathbf{p}=[x,y,z,i,f^{(64)}]$ via this mapping ($\Pi: \mathbb{R}^2 \mapsto \mathbb{R}^3$), where $i$ denotes its intensity. 

Through the FPR module, the 2D semantic feature points extracted from the range view are seamlessly integrated with the 3D point clouds, effectively enabling the exploitation of the semantic information for further 3D object detection. Essentially, this module harmonizes the 2D semantic feature points derived from the range view with the 3D point cloud. These points are subsequently re-transformed into their 3D coordinates and further encoded by 3D CNNs. Since the 3D CNNs can extract 3D surface texture information, they mitigate the inherent limitations of the range-view representation. Overall, this plug-and-play module demonstrates remarkable efficiency and efficacy compared to point-wise processing methods like IA-SSD\cite{ia-ssd} for the 3D object detection task, as validated through Table~\ref{tab:effect_of_R2}. 

\textbf{R\textsuperscript{2} Module}. Our proposed R\textsuperscript{2}M combines the Range-view Feature Extraction and FPR components to create a unified structure. This structure enhances the utilization of range-view feature representation for 3D object detection. On the one hand, leveraging the dense semantic information in range images compared to 3D sparse points, the R\textsuperscript{2}M exploits these dense 2D features and integrates geometric information by expanding the receptive field in 3D space through HD Meta Kernel. This expansion significantly contributes to enhancing detection accuracy. On the other hand, the FPR module smoothly converts all redeemed feature points into 3D space, enabling efficient processing and seamless integration of the R\textsuperscript{2}M into existing LiDAR-based 3D object detectors. As illustrated in Table~\ref{tab:effect_of_R2}, the R\textsuperscript{2}M enhances the input-level 3D points with range-view semantic features, leading to improved detection performance across various 3D detectors.



\begin{figure}[t]
    \centering
    \includegraphics[width=0.8\linewidth]{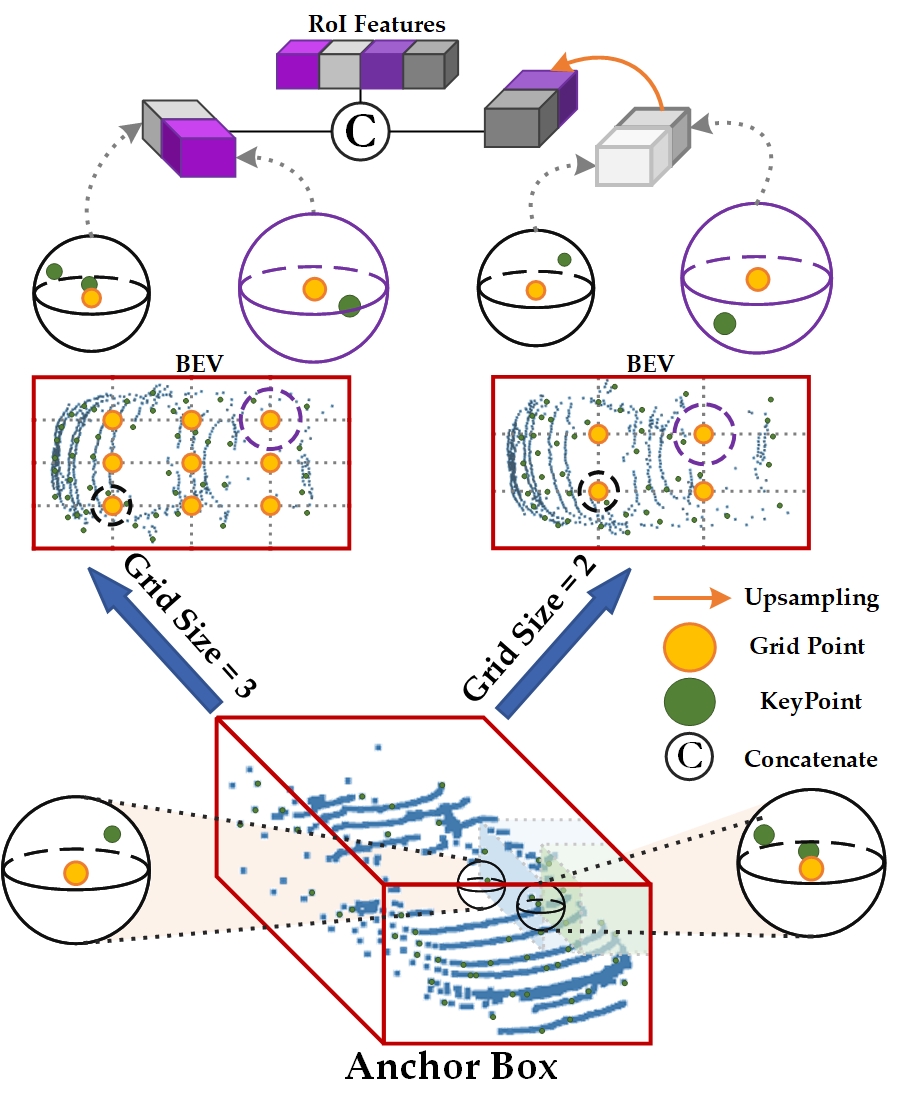}
    \vspace{-5mm}
    \caption{Illustration of the S-Grid RoI Pooling module. A slight change in the sampling site (Grid Point) causes considerably different feature distribution of sampled keypoints.}
    \label{fig:SGRoIPooling}
\end{figure}


\subsection{Stage II: 3D Voxel-based Feature Extraction and Proposal Generation} \label{sec:RPFE}
Previous 3D detectors \cite{pvrcnn, voxelrcnn, second,submanifold,voxelnet} frequently utilize voxel-based 3D sparse convolution to transform point clouds into sparse 3D volumes and embeds them with extracted voxel-wise features. The resulting tensors are subsequently layered along the Z axis to generate feature maps in the bird's-eye view. For its efficacy and precision, we employ it as the 3D backbone of our framework for 3D texture feature capture and 3D proposal generation from the redeemed feature points. More detailed descriptions on Stage \Romannum{2} are provided in sup. material.

With the seamless connection of range-view and voxel-wise processing, R\textsuperscript{2}Det extracts versatile features from flexible viewpoints, offering both complementary 2D semantic and 3D texture information. In contrast to range-view-based methods \cite{rangedet,rangeioudet,rsn,lasernet,RCD,rangercnn}, which excel in 2D semantic features but lack 3D surface texture information due to limitations of range images, our R\textsuperscript{2}Det addresses and mitigates this issue by employing subsequent 3D CNNs to restore 3D texture information from the redeemed feature points in the range view. Unlike multi-view approaches \cite{MVF, MV3D, pillarod, h23drcnn}, which typically require fusing 2D views of point clouds, the FPR strategy facilitates feature extraction across multiple dimensions directly from the 3D point cloud naturally generated from the range image, eliminating the need for fusion stages.


\subsection{Stage III: Box Refinement} \label{sec:boxrefine}
Since redeemed feature points are separated into regular voxels globally in the prior modules, and candidate anchors are positioned in the bird's-eye view, this section identifies local features to facilitate the refinement of candidate boxes. This section is comprised of three subsections: the global scenario representation, the local RoI feature extraction, and the box refinement.
\begin{table*}[ht]
    \centering
    \setlength{\tabcolsep}{4pt} 
    \renewcommand{\arraystretch}{0.95} 
    \begin{tabularx}{\textwidth}{cc *{6}{Y}}
    \toprule
        \multirow{2}{*}{Method}&\multirow{2}{*}{Input View}&\multicolumn{3}{c}{Car AP$_{\textbf{3D}}$(\%)}&\multicolumn{3}{c}{Car AP$_{\textbf{BEV}}$(\%)}\\
&&Easy&Moderate&Hard&Easy&Moderate&Hard\\\hline

CAT-Det\cite{cat}&MV&89.87&81.32&76.68&92.59&{90.07}&85.82\\
EPNet++\cite{epnet++}&MV&\textbf{91.37}&81.96&76.71&\textbf{95.41}&89.00&85.73\\
MVMM\cite{mvmm}&MV&87.59&78.87&73.78&92.17&88.70&85.47\\ 
PVT-SSD\cite{pvtssd}&MV&90.65 &82.29 &76.85& 95.23 & \textbf{91.63} &\textbf{86.43} \\
DVF-PV\cite{dvf} & MV & 90.99 &\textbf{82.40} &\textbf{77.37}&-&-&-\\ \hline \hline
Voxel R-CNN \cite{voxelrcnn}&BEV&\textbf{90.90} &81.62&77.06&\textbf{94.85}&88.83&86.13\\
VoxSeT\cite{voxset}&BEV&88.53&82.06&77.46&92.70&89.07&86.29 \\
BSAODet(PV-RCNN++)\cite{bsaodet}&BEV&88.66&81.95&77.40&	92.66 &	88.90 &	86.23\\
OcTr\cite{octr} &BEV&90.88&\textbf{82.64}&\textbf{77.77}&93.08&\textbf{89.56} &\textbf{86.74} \\ 
GD-MAE\cite{gd-mae}&BEV&88.14&79.03&	73.55&94.22&88.82&83.54 \\ \hline \hline
PointRCNN\cite{pointrcnn}&PV&86.96&75.64&70.70&92.13 &87.39&82.72\\
3DSSD\cite{3dssd}&PV&88.36 & 79.57& 74.55 & 92.66  & 89.02 & 85.86 \\
IA-SSD(multi) \cite{ia-ssd}&PV&88.34&80.13&75.04&92.79&89.33&84.35\\
SASA \cite{Chen2022SASASS}&PV&\textbf{88.76}&\textbf{82.16}&\textbf{77.16}&\textbf{92.87}&\textbf{89.51}&\textbf{86.35}\\
DTSSD\cite{dtssd}&PV&88.62&79.94&74.91&92.77&89.18 &86.03 \\ \hline \hline
RangeRCNN \cite{rangercnn}&RV&88.47&81.33&77.09&92.15&88.40&85.74\\
RCD\cite{RCD}&RV&70.54 &60.56 &55.58 &82.26&75.83 &69.61\\
RangeIoUDet\cite{rangeioudet}& RV&88.60&79.80&76.76&92.28&88.59&85.83 \\
RangeDet \cite{rangedet}&RV&85.41&77.36&72.60&90.93&	87.67 &	82.92 \\
\rowcolor[HTML]{DAD8D8} 
R\textsuperscript{2}Det (ours)&RV&\textbf{90.93}&\textbf{82.42}&\textbf{77.84}&\textbf{93.02}&\textbf{91.17}&\textbf{86.62}\\ \bottomrule
    \end{tabularx}
    \vspace{-3mm}
     \caption{Comparison on the  KITTI online \textit{test} server. AP with an IoU threshold of 0.7 for \textit{Cars} and 40 recall positions are conducted to evaluate the results. $80\%$ of the total \textit{train+val} data is used for training. MV, BEV, PV and RV refer to methods with input from multiple views, bird's-eye view, point view and range view.}
     \label{tab:kitti-test}
\end{table*}
\textbf{Global Scenario Representation.} As the 3D voxel CNNs serve as the 3D backbone of our model, a straightforward option for this step is directly implementing the Voxel Set Abstraction (VSA) paradigm proposed in \cite{pvrcnn}. In the VSA, the PointNet block\cite{pointnet} transforms each level of voxel-wise features and concatenates them together. However, this procedure involves processing and storing large amounts of data, resulting in demanding computational requirements and longer training time.

Given the limitations of VSA, we opt to apply the Furthest Point Sampling \cite{pointnet++} algorithm to downsample $\mathcal{P}$ into a set of representative points $\mathcal{R}=\{\mathbf{p_j} |\,\, \forall \mathbf{p_j} \in \mathcal{P}, \mathbf{j} = [0,C-1]\}$, where $C$ specifies the size of the point set. Subsequently, the redeemed feature points undergo another round of feature extraction using the PointNet block within the PV. These simple operations aim to enrich contextual data, facilitating subsequent proposals refinement module. Compared with the VSA module, we provide a direct yet more appropriate solution for processing the redeemed feature points.

\textbf{Local RoI Feature Extraction.} The RoI-grid pooling algorithm in PV R-CNN inspires us to aggregate RoI features from each grid point using surrounding keypoints. However, the keypoints within a certain radius have a sensitive structure that highly depends on their sampling position. Even a small adjustment to the sampling site can lead to very different feature aggregation, as shown in Figure~\ref{fig:SGRoIPooling}.

As shown in Figure~\ref{fig:SGRoIPooling}, feature distribution is susceptible to (1) sampling radii and (2) grid sizes. Consequently, we partition the candidate box into two grid sizes of 3 and 2, and then apply RoI-grid pooling with distinct sampling radii in each branch. The resulting refined features are then concatenated to form RoI features. Particularly, an upsampling module is implemented in the rough partition (Grid Size $=2$) branch to achieve uniform scaling prior to concatenation. Compared with the RoI-grid Pooling strategy, the proposed S-Grid RoI Pooling leverages more comprehensive contextual information for each proposal.

\textbf{Detect Head.} After RoI pooling, the detect head predicts a confidence score related to Intersection over Union (IoU) and conducts additional regression on the box coordinates for each region proposal. The refinement network consists of a two-layer Multilayer Perceptron (MLP) according to \cite{pvrcnn} with separate branches for confidence prediction and box refinement.

\subsection{Training Loss}
The training process of R\textsuperscript{2}Det is end-to-end optimized, whose total training loss can be calculated as:
\begin{equation}
    \mathcal{L}_{T} = \mathcal{L}_{rpn} + \mathcal{L}_{rea} +\mathcal{L}_{ref}
    \label{eqn:totalloss}
\end{equation}
where $\mathcal{L}_{rpn}$ represents the loss of the proposal generation module referring to \cite{second}, $\mathcal{L}_{rea}$ and $\mathcal{L}_{ref}$ denote the training objective of the weight reassignment module and the training loss of the box refinement module according to \cite{pvrcnn}. For more detailed information on training losses, please refer to the sup. file.

\section{Experiments}

\begin{table}[t]
\centering
    \setlength{\tabcolsep}{3pt} 
    \renewcommand{\arraystretch}{0.6} 
\resizebox{\columnwidth}{!}{%
\begin{tabular}{cccc}
\toprule
\multirow{2}{*}{Method} & \multicolumn{3}{c}{Car AP$_{\textbf{3D}}$(\%)} \\
& Easy & Mod. & Hard \\ \hline
\multicolumn{2}{l}{\textbf{Multiple View Input:}} & \textbf{} & \textbf{} \\
H\textsuperscript{2}3D R-CNN\cite{h23drcnn}  & 89.63 & 85.20 & 79.08 \\ 
CAT-Det\cite{cat}&{90.12}&81.46&{79.15} \\ 
EPNet++\cite{epnet++}&\textbf{92.51}&\textbf{83.17}&\textbf{82.27} \\ \hline \hline
\multicolumn{2}{l}{\textbf{Single View Input:}} & \textbf{} & \textbf{} \\
PointPillars\cite{pointpillars} & 86.62 & 76.06 & 68.91 \\
PV-RCNN\cite{pvrcnn} & 89.35 & 83.69 & 78.70 \\
CIA-SSD\cite{ciassd} & \textbf{90.04} & 79.81& 78.80 \\
VoTr-TSD\cite{votr}&89.04&84.04&78.68\\
Voxel R-CNN\cite{voxelrcnn}  & 89.41 & 84.52 & 78.93 \\
IA-SSD\cite{ia-ssd}  & - & 79.57& - \\
VoxSeT\cite{voxset}&88.45&78.48&77.07\\
DTSSD\cite{dtssd}&-&79.39&-\\
OcTr\cite{octr}&88.43 &78.57 &77.16\\ 
\rowcolor[HTML]{DAD8D8} 
R\textsuperscript{2}Det (ours) & 89.75& \textbf{85.78}& \textbf{79.08}\\ \bottomrule
\end{tabular}
}
\vspace{-3mm}
\caption{Comparison on the KITTI \textit{val} split with AP derived from 11 recall positions for \textit{Car} class.}
\label{tab:kitti-val}
\end{table}

We conduct experiments on both the KITTI dataset and the Waymo Open Dataset. For the KITTI dataset, we follow the official \textit{train}/\textit{val} split. More detailed configuration of the pipeline is provided in the sup. file.

\subsection{Detection Results on KITTI Dataset}


As shown in Table \ref{tab:kitti-test}, R\textsuperscript{2}Det achieves the state-of-the-art (SOTA) performance on the KITTI \textit{test} set thanks to R\textsuperscript{2}M utilizing the range-view representation to its full potential. 
It outperforms all range-view-based methods at all difficulty levels. Generally, R\textsuperscript{2}Det achieves 77.84\% on the hard level of \textit{Car} in AP$_{\textbf{3D}}$, beating other MV, BEV, and PV-based algorithms. 

For a comprehensive comparison, we also provide the performance of the KITTI \textit{val} split with mAP derived from 11 recall positions. According to Table \ref{tab:kitti-val}, our R\textsuperscript{2}Det has the strongest effect on the \textit{val} split at the moderate and hard difficulty levels. In summary, our proposed R\textsuperscript{2}Det is competitive to MV, BEV, and PV-based methods and achieves SOTA performance on range-view-based 3D object detection, as demonstrated by the results on both the \textit{test} set and \textit{val} split of the KITTI.


%
\subsection{Detection Results on Waymo Open Dataset}

\begin{table}[t]
\centering
\setlength{\tabcolsep}{1.5pt} 
\renewcommand{\arraystretch}{0.6} 
\begin{tabular}{cccc}
\toprule
 Method& \begin{tabular}[c]{@{}c@{}}Veh. (\%)\\ mAP/mAPH\end{tabular} & \begin{tabular}[c]{@{}c@{}}Ped.(\%)\\ mAP/mAPH\end{tabular} \\ \hline
 \textbf{\textit{Level 1:}}&&\\
 $\dag$PointPillars\cite{pointpillars}& 56.6/-& 59.3/-\\
 *PV-RCNN \cite{pvrcnn}& 74.1/73.4 & 62.7/52.7\\
 RSN \textit{label}\tablefootnote{\label{footnote}The RSN model for the Vehicle label is RSN Car\_1f, for the Pedestrian label is RSN Ped\_1f} S\_1f \cite{rsn}& 70.5/70.0 & {74.8}/{69.6}\\
 RSN\textit{label}\footref{footnote}L\_1f \cite{rsn}&75.1/74.6& {77.8}/{72.7}\\
IA-SSD \cite{ia-ssd}& 70.5/69.7& 69.4/58.5\\
VoxSeT\cite{voxset}& 74.5/74.0& 72.5/65.4\\
GD-MAE\cite{gd-mae}&77.3/76.8&80.3/72.4\\
VoxelNeXt\cite{voxenext}&78.2/77.7 &\textbf{81.5}/\textbf{76.3} \\
BSAODet\cite{bsaodet}&78.3/77.6&75.9/64.4\\
 R\textsuperscript{2}Det$_{0.25}$(ours)& {75.4}/{74.8}& 73.4/63.7\\
 R\textsuperscript{2}Det(ours)& \textbf{78.4}/\textbf{77.9}&77.9/75.0\\ \hline
  \textbf{\textit{Level 2:}}&&\\
 *PV-RCNN\cite{pvrcnn} &65.0/64.3&53.8/45.1\\
 RSN \textit{label}\footref{footnote}S\_1f \cite{rsn}& 63.0/62.6& {65.4}/{60.7}\\
 IA-SSD \cite{ia-ssd}& 61.6/ 60.8&60.3/50.7\\
 VoxSeT\cite{voxset}&66.0/65.6&-/-\\
 GD-MAE\cite{gd-mae}&68.7/68.3&\textbf{72.8}/65.5\\
 VoxelNeXt\cite{voxenext}& 69.7/69.2 &72.2/\textbf{65.9} \\
 BSAODet\cite{bsaodet}&69.5/68.9&66.8/56.4\\
 R\textsuperscript{2}Det$_{0.25}$(ours)& {67.3}/{66.7}& 64.3/55.6\\ 
 R\textsuperscript{2}Det(ours)& \textbf{70.2}/\textbf{69.5}&67.0/63.2\\ \bottomrule
\end{tabular}
\vspace{-3mm}
\caption{Comparison on the Waymo Open Dataset with 202 validation sequences for the 3D vehicle (IoU = 0.7), pedestrian
(IoU = 0.5). Subscript 0.25: using 25\% training data. $\dag$: re-implemented by \cite{MVF}. *: provided by \cite{ia-ssd}}
\label{tab:waymo-difficulty}
\end{table}

\begin{table}[h]
\centering
\setlength{\tabcolsep}{1.5pt} 
\renewcommand{\arraystretch}{0.7} 
\resizebox{\columnwidth}{!}{%
\begin{tabular}{ccccc}
\toprule
& \multicolumn{4}{c}{mAP$_{\textbf{3D}}$(\%)}\\
\multirow{-2}{*}{Method} & Overall& \textless{}30m& 30-50m&$\geqslant$50m\\ \hline
LaserNet\cite{lasernet}& 52.11& 70.94& 52.91& 29.62\\
RCD\cite{RCD}& 66.39& 86.59& 65.64& 40.00\\
RSN CarS\_1f\cite{rsn}& 70.50& 90.80& 67.80& 45.40\\
RangeDet\cite{rangedet}& 72.85& 87.96& 69.03& 48.88\\
R\textsuperscript{2}Det$_{0.25}$(ours)& {75.36} & {92.00} & {73.17} & {52.15} \\ 
R\textsuperscript{2}Det(ours)& \textbf{78.38} & \textbf{92.67} & \textbf{76.85} & \textbf{56.67} \\
\rowcolor[HTML]{96FFFB} 
\textit{Improvement}&\textit{+5.53}&\textit{+1.87}&\textit{+7.82}&\textit{+7.79}\\ \bottomrule
\end{tabular}
}
\vspace{-3mm}
\caption{Comparison with other methods taking range view as input on the Waymo Open Dataset with 202 validation sequences.}
\label{tab:waymo-improvements}
\end{table}
\begin{table}[ht]
\centering
    \setlength{\tabcolsep}{3pt} 
    \renewcommand{\arraystretch}{0.8} 
\begin{tabular}{cccccc}
\toprule
\multirow{3}{*}{Model} & \multicolumn{4}{c}{Range-view Feature Extraction} & \multirow{3}{*}{\begin{tabular}[c]{@{}c@{}}AP$_{\textbf{3D}}$ \\ (\%)\end{tabular}} \\
 & Layer 1 & \multicolumn{3}{c}{Layer 2}&  \\
 & RB & RB & MK & HD-MK &  \\ \hline
Baseline & \checkmark &  &  &  & 81.33 \\
R\textsuperscript{2}Det& \checkmark & \checkmark &  &  & 81.98\\
R\textsuperscript{2}Det& \checkmark &  & \checkmark &  & 83.19\\
R\textsuperscript{2}Det& \checkmark &  & & \checkmark  & \textbf{85.71}\\ \bottomrule
\end{tabular}
\vspace{-3mm}
\caption{Effect of the HD Meta Kernel encoding strategy on the KITTI \textit{val} split. RB indicates the residual block~\cite{basicblock}, MK indicates the \textit{Meta-kernel} convolution layer.}
\label{tab:ablationHDmetakernel}
\end{table}
As the Waymo Open Dataset (WOD) is the unique dataset providing range images, we further examine the performance of R\textsuperscript{2}Det on this challenging large-scale dataset. 


Among the methods listed in Table~\ref{tab:waymo-difficulty}, R\textsuperscript{2}Det achieves the best results in two difficulty levels, with the highest mAP/mAPH for the \textit{Vehicle} class and the second best mAPH for the \textit{Pedestrian} class on Level 1 difficulty level. We can observe that R\textsuperscript{2}Det is highly competitive with state-of-the-art methods on the Waymo Open Dataset, achieving the best performance on both difficulty levels for the \textit{Vehicle} class with an mAP of $78.4\%$ and $70.2\%$, respectively. 

Table~\ref{tab:waymo-improvements} validates the importance of addressing the issue of 3D surface texture loss, which is ignored by all existing range-view-based approaches. By recovering the 3D surface shape of objects through our R\textsuperscript{2}M, R\textsuperscript{2}Det outperforms existing range-view-based methods by a substantial margin of $+5.53\%$ overall. Especially, R\textsuperscript{2}Det achieves absolute improvements of $+7.82\%$ in the $30-50m$ distance range and $+7.79\%$ in distances exceeding $50m$.

In conclusion, our results demonstrate the superiority of R\textsuperscript{2}Det for range-view-based 3D object detection on the WOD and validate our perspective outlined in Section 1.

\begin{figure*}[t!]
    \centering
    \includegraphics[width=0.99\linewidth]{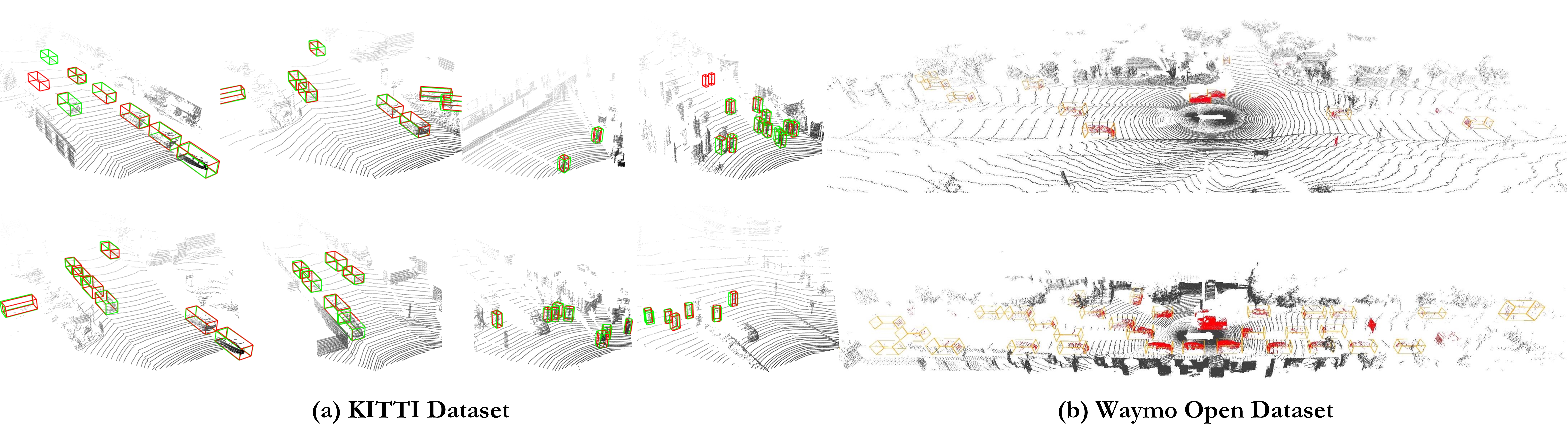}
    \vspace{-5mm}
    \caption{Quantitative results on KITTI and WOD datasets. (a) The \textcolor{red}{red} bounding boxes are predictions, and the \textcolor{green}{green} ones are ground truths. (b) The \textcolor{red}{red} points are ground truths, and the \textcolor{orange}{orange} boxes are the predictions. More results refer to the sup. file.}
    \label{fig_vis}
\end{figure*}

\subsection{Ablation Study}
We conduct extensive ablation experiments to assess separate portions of our proposed scheme. All models are trained on the KITTI \textit{train} split and evaluated on the KITTI \textit{val} split. All AP scores are calculated by $40$ recall positions.

\textbf{The effectiveness of R\textsuperscript{2} Module.} Table~\ref{tab:effect_of_R2} indicates the substantial enhancements that our plug-and-play R\textsuperscript{2}M bring to the existing 3D object detection methods. Across various categories, including point-voxel-based, voxel-based, and point-based approaches, our R\textsuperscript{2}M consistently demonstrates significant improvements of $1.39\%$, $1.67\%$ and $1.97\%$ mAP improvement on easy, moderate and hard difficulty level, respectively.

\textbf{The effectiveness of HD Meta Kernel.} Table~\ref{tab:ablationHDmetakernel} shows the impact of different range-view feature extraction strategies, which contain 2 layers. The 1st layer is a residual block (RB) of BasicBlock~\cite{basicblock}, while the 2nd layer can be one of the RB, Meta Kernel, or HD Meta Kernel. The baseline model, which uses only one layer of BasicBlock for feature encoding, is presented in the first row. The substantial enhancement of $+3.73\%$ between the second and last rows highlights the impact of this module. Similarly, the $+2.52\%$ increment between the last two rows indicates the performance boost achieved by expanding the 3D receptive field through the HD Meta Kernel.

\begin{table}[h]
\centering
    \setlength{\tabcolsep}{6pt} 
    \renewcommand{\arraystretch}{0.7} 
\begin{tabular}{cccccc}
\toprule
\multicolumn{3}{c}{Range-view Feature Extraction} & \multicolumn{2}{c}{2D to 3D} & \multirow{2}{*}{\begin{tabular}[c]{@{}c@{}}AP$_{\textbf{3D}}$\\ (\%)\end{tabular}} \\
RB & HD-MK & U-Net+Seg. & Fore. & FPR &  \\ \midrule
\checkmark&  &  &  & \checkmark &  81.33\\
&  & \checkmark & \checkmark &  &82.78  \\
\checkmark&  &  \checkmark&  & \checkmark & 84.41 \\
\checkmark& \checkmark &  &  & \checkmark & \textbf{85.71} \\ \bottomrule
\end{tabular}
\vspace{-3mm}
\caption{Effect of the Feature Points Redemption compared with \cite{rsn}. FPR is our Feature Points Redemption strategy. The U-Net+Seg. refers to \cite{rsn}, and Fore. stands for only foreground points transformation.}
\label{tab:ablation_FPR}
\end{table}

\textbf{The effectiveness of HD-MK and FPR in maintaining the continuous 3D surface texture information.} We substantiate the importance of continuous 3D surface texture information for 3D object detection presented in Table~\ref{tab:ablation_FPR}. We compare our continuous 2D feature extraction (HD-MK) method with whole scene Feature Point Redemption (FPR) against the binary-segmentation method with the foreground point transformation in \cite{rsn}. This method segments foreground points and performs 2D to 3D transformation, disconnecting 3D surface textures between foreground and background. The difference of $1.63\%$ between the second and third rows supports the notion that the absence of continuous 3D surface texture can result in suboptimal performance. Furthermore, the difference of $1.30\%$ between the last two rows indicates our 2D feature extraction method outperforms simple binary segmentation.


\textbf{The effectiveness of S-Grid RoI Pooling.} Table~\ref{tab:effect_of_pooling} shows the effect of our proposed S-Grid RoI Pooling on two different models, i.e., \cite{pvrcnn} and our R\textsuperscript{2}Det. We compare each model's performance with RoI-grid pooling and S-Grid RoI pooling, respectively. The results demonstrate that using S-Grid RoI pooling consistently outperforms RoI-grid pooling for both models. This validates that enriching features with various sampling sites is effective in improving the performance of the models. Furthermore, the results suggest that keypoints within a certain radius have a sensitive structure that highly depends on their sampling position, as stated in the Methodology.


\begin{table}[t]
\centering
    \setlength{\tabcolsep}{3.0pt} 
    \renewcommand{\arraystretch}{0.8} 
\begin{tabular}{cccc}
\toprule
 & \multicolumn{3}{c}{AP$_{\textbf{3D}}$(\%)} \\
\multirow{-2}{*}{Method} & Easy & Mod. & Hard \\ \hline
PV-RCNN w RoI-grid Pooling & 91.31 & 82.66 & 80.08 \\
PV-RCNN w RoI-grid Pyramid & 92.13 &83.13 &82.40\\
PV-RCNN w S-Grid RoI Pooling & 92.44 & 84.77 & 82.48 \\
\rowcolor[HTML]{96FFFB} 
\textit{Improvement} & \textit{+0.31} & \textit{+1.64} & \textit{+0.08}\\ \hline
R\textsuperscript{2}Det w RoI-grid Pooling & 89.42 & 83.38 & 80.20 \\
R\textsuperscript{2}Det w RoI-grid Pyramid&91.19& 83.82 &82.65\\
R\textsuperscript{2}Det w S-Grid RoI Pooling & \textbf{92.82} & \textbf{85.71} & \textbf{83.36} \\
\rowcolor[HTML]{96FFFB} 
\textit{Improvement} & \textit{+1.63} & \textit{+1.89} &\textit{+0.71} \\ \bottomrule
\end{tabular}
\vspace{-3mm}
\caption{Effect of the S-Grid RoI Pooling on the KITTI \textit{val} split with AP calculated by 40 recall positions. RoI-grid Pyramid refers to \cite{pyramidrcnn}}
\label{tab:effect_of_pooling}
\end{table}

\section{Conclusion}
We present R\textsuperscript{2}Det, a flexible viewpoint solution for 3D object detection that uses the range view of point clouds as input. Our approach generates range-viewed feature maps using the HD Meta Kernel and uses the Feature Points Redemption module to convert range-view feature points to point view. Proposals are generated from bird's-eye-view features and refined after S-Grid RoI pooling. Despite using a single-view representation, our approach incorporates information from multiple viewpoints through the Feature Points Redemption module. R\textsuperscript{2}Det outperforms all range-view-based approaches and achieves impressive results on both the KITTI dataset and the Waymo Open Dataset. Our findings highlight the importance of addressing 3D surface texture loss in range-view representation and suggest R\textsuperscript{2}Det offers an elegant range-view exploration solution for 3D object detection.

\bibliography{egbib}

\end{document}